\documentclass[conference]{IEEEtran}
\IEEEoverridecommandlockouts
\usepackage{cite}
\usepackage{amsmath,amssymb,amsfonts}
\usepackage{algorithmic}
\usepackage{graphicx}
\usepackage{textcomp}
\usepackage{xcolor}
\usepackage[caption=false]{subfig}
\usepackage{booktabs}

\def\BibTeX{{\rm B\kern-.05em{\sc i\kern-.025em b}\kern-.08em
    T\kern-.1667em\lower.7ex\hbox{E}\kern-.125emX}}

\usepackage{acronym}
\newacro{AI}{Artificial Intelligence}
\newacro{XAI}{Explainable AI}
\newacro{NLP}{Natural Language Processing}
\newacro{ML}{Machine Learning}
\newacro{DL}{Deep Learning}
\begin{document}

\title{Enhancing the Fairness and Performance of Edge Cameras with Explainable AI
}

\author{\IEEEauthorblockN{
Truong Thanh Hung Nguyen\IEEEauthorrefmark{2}\IEEEauthorrefmark{3},
Vo Thanh Khang Nguyen\IEEEauthorrefmark{3},
Quoc Hung Cao\IEEEauthorrefmark{3},\\ 
Van Binh Truong\IEEEauthorrefmark{3},
Quoc Khanh Nguyen\IEEEauthorrefmark{3},
Hung Cao\IEEEauthorrefmark{2}}
\IEEEauthorblockA{\IEEEauthorrefmark{3}Quy Nhon AI, FPT Software, Vietnam
\IEEEauthorrefmark{2}Analytics Everywhere Lab, University of New Brunswick, Canada\\
Email: \{hungntt, khangnvt1, hungcq3, binhtv8, khanhnq33\}@fpt.com, hcao3@unb.ca}
}

\maketitle

\begin{abstract}
The rising use of \ac{AI} in human detection on Edge camera systems has led to accurate but complex models, challenging to interpret and debug. Our research presents a diagnostic method using XAI for model debugging, with expert-driven problem identification and solution creation. Validated on the Bytetrack model in a real-world office Edge network, we found the training dataset as the main bias source and suggested model augmentation as a solution. Our approach helps identify model biases, essential for achieving fair and trustworthy models.
\end{abstract}
\begin{IEEEkeywords}
Explainable AI, Edge Camera
\end{IEEEkeywords}

\section{Introduction}
\begin{figure}[htp]
    \centering
    \subfloat[\label{subfig:camera_cover_body}]{\includegraphics[width=.4\linewidth]{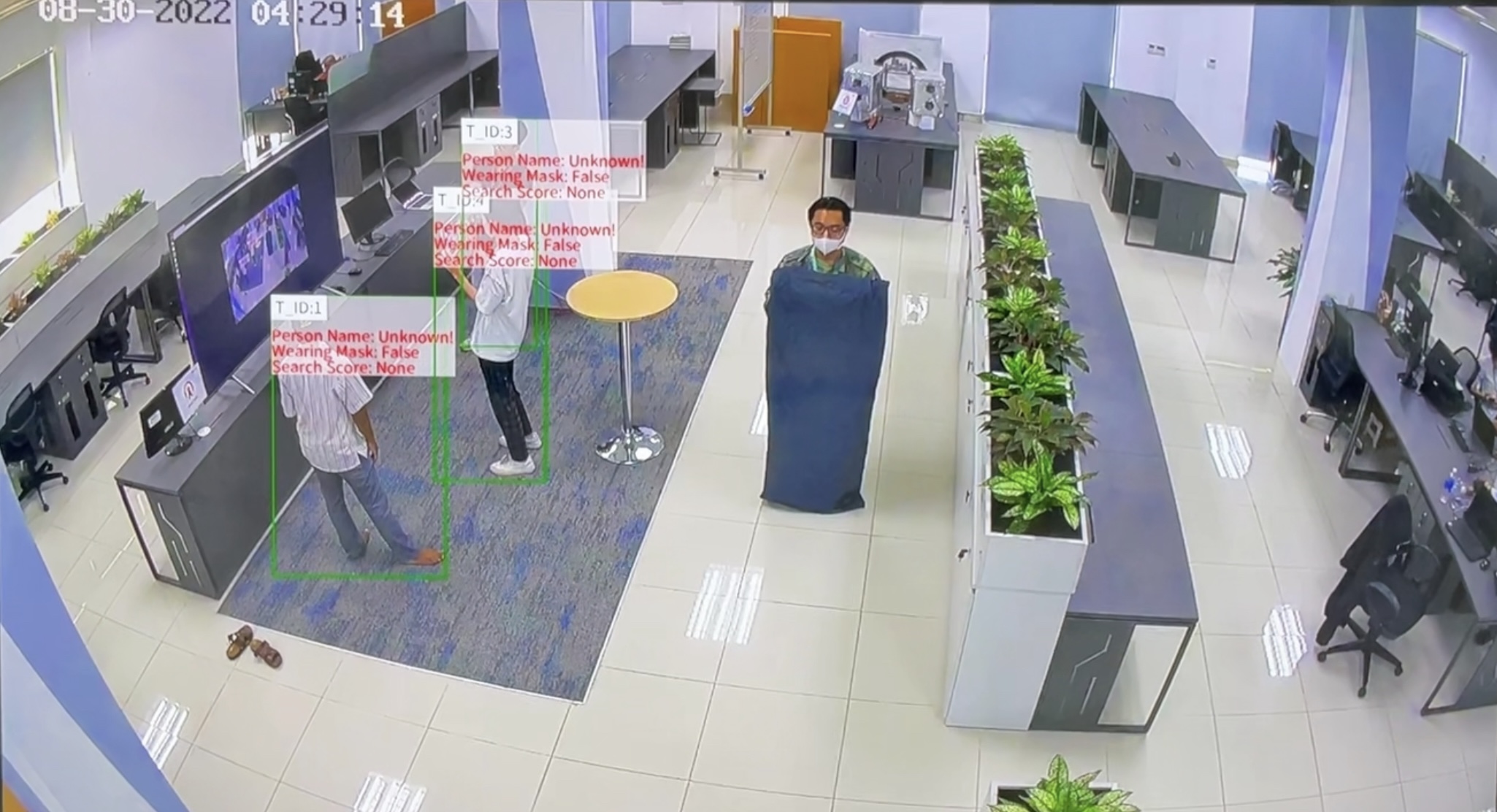}}
    \subfloat[\label{subfig:disabled}]{\includegraphics[width=.4\linewidth]{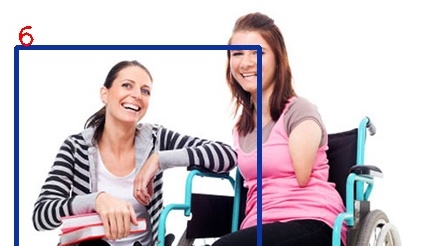}}
    \caption{(a) A security camera on the ceiling of an office can detect ordinary people (green boxes), but not people who cover their bodies with a cloth. (b) The Bytetrack model cannot detect the disabled woman but still detect the other, who is not disabled.}
    \label{fig:camera_cover_body}
    \vspace{-10pt}
\end{figure}

Human detection through security cameras, a pivotal \ac{AI} task, employs AI models like YOLO and its YOLOX variant for alerts, such as falls and intrusions. Specifically, Bytetrack, based on YOLOX, excels in multi-object tracking~\cite{yolox, bytetrack}. Yet, it struggles in detecting obscured or disabled individuals (Fig.~\ref{subfig:camera_cover_body}, Fig.~\ref{subfig:disabled}). Given their black-box nature, these models pose debugging challenges. Though XAI aids debugging in tabular and text data~\cite{auditing_debugging}, its use in image data is less explored. Hence, our paper introduces an XAI-driven framework to debug human detection models in security cameras. The approach leverages experts for diagnosing problems and proposing solutions, with potential wider relevance to object detection and classification.

\section{Related Work}
\subsection{Human Detection}
Human detection identifies humans in images or videos and has evolved with various methods. Deep Learning (DL) brought forward models that address challenges like object size and illumination differences. Capitalizing on YOLOX's~\cite{yolox} success, Bytetrack~\cite{bytetrack} was designed for human detection, leveraging YOLOX for detection and Byte for post-processing.

\subsection{Explainable AI}
AI's integration into real-world scenarios has led to multiple \ac{XAI} strategies: perturbation-based, backpropagation-based, and example-based.
Perturbation techniques, such as D-RISE~\cite{drise}, which work independently of model design, perturb input images, then analyze predictions to gauge pixel or superpixel influence on outcomes. While widely applicable, their computational demand can be limiting.
Backpropagation methods delve into model architecture to fetch explanatory data. Recognized techniques include GradCAM~\cite{gradcam}, SeCAM~\cite{secam}.
Example-based methods, like Influence Function~\cite{influence_function}, explain using training data samples to ascertain their effects on predictions.
While XAI's application to object detection is complex due to the intricate models, some methods, such as D-RISE~\cite{drise}, D-CLOSE~\cite{truong2023towards}, and G-CAME~\cite{nguyen2023g}, are adaptations from classification for object detection.

\subsection{Debugging Model Framework with XAI}
Many studies utilize XAI methods~\cite{xai_akamedic}, primarily answering, \textit{``Why does the model predict this?''} Yet, the follow-up, \textit{``How can explanations improve the model?''} requires using XAI to better the AI system. No research has yet outlined a framework for debugging human detection models. This paper, therefore, introduces such a framework, leveraging XAI to pinpoint issues and improve model fairness and efficacy.


\section{Methodology}
\begin{figure}[h]
          \centering
          \includegraphics[width=\linewidth]{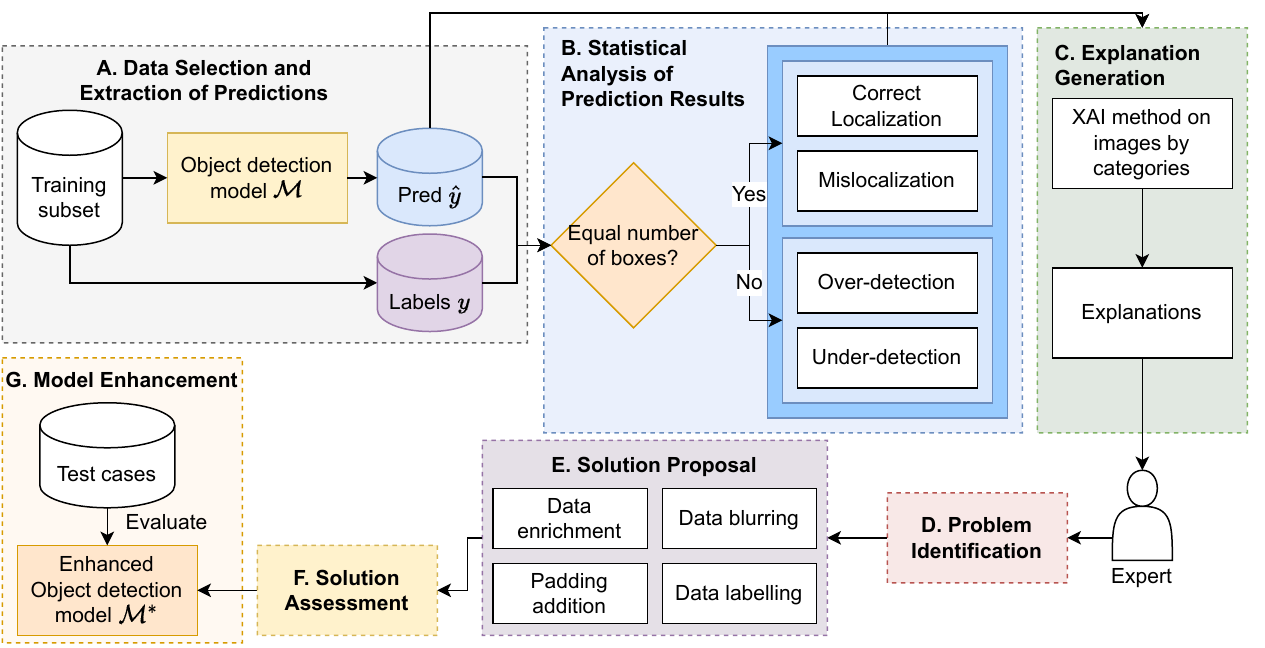}
        \caption{The Debugging Framework for Human Detection Models}
        \label{fig: xai process}
        \vspace{-10pt}
\end{figure}

We present a structured debugging model framework shown in Fig.~\ref{fig: xai process}, with seven sequential stages. Each stage relies on the results of its predecessor. Where multiple methods or assumptions exist per stage, we offer strategy selection guidelines. In this framework, XAI aids experts in identifying core model issues and suggesting performance-enhancing solutions.

\subsection{Data Selection and Extraction of Predictions}~\label{s:data_selection}
Our framework starts by selecting a training dataset subset for model enhancement, addressing potential dataset concerns. Public datasets like CrowdHuman~\cite{crowdhuman}, used in Bytetrack training, can face data poisoning~\cite{data_poisoning}, affecting data quality and model results.
Error detection in the model or dataset is optimized using random testing~\cite{mayer2006empirical}, which randomly picks data for testing, spotting major flaws without full dataset checks.
Based on the idea that small samples can be indicative, we use statistical sampling heuristics to set an optimal sample size, which should not surpass 10\% of the full dataset or 1000 samples, ensuring a meaningful and efficient subset~\cite{understanding_rule_of_thumb}.
After selecting the data subset, it's fed into the model to generate predictions. These are then analyzed against the ground truth, helping gauge model metrics like accuracy, precision, and areas needing enhancement.

\subsection{Statistical Analysis of Prediction Results}\label{s:analysis}
After obtaining predictions, they are categorized by comparing them with the ground-truth. This classification is guided by experts and, in our human detection context, results in four categories.
Initially, dataset categorization relies on whether the model's predicted count aligns with the ground truth. Images are labeled as ``Under-detection'' if the model detects fewer people, and ``Over-detection'' if it detects more.
If the model's count matches the ground truth, detection quality is evaluated by comparing model-detected boxes with ground truth boxes using Intersection over Union (IoU) values. Images with all box pairs having $\text{IoU}\geq0.5$ are deemed ``Correct Localization'', while others are ``Mislocalization''.

This process organizes the dataset based on prediction results, with three categories signaling potential model enhancements. The next stage delves deeper into error sources, laying the groundwork to boost the model's precision in detecting people within images.

\subsection{Explanation Generation}
In this phase, we use XAI methods to explain each image category. Given that D-RISE~\cite{drise} is adaptable to diverse models without needing their architecture details and offers explanations for ground truth boxes (enabling comparison with model-detected boxes), we opt for D-RISE in human detection. These explanations assist experts in identifying the root of incorrect predictions in the following stage.

\subsection{Problem Identification}~\label{s:problem}
Using the XAI results from the prior phase, experts analyze each category presented in the statistical analysis (Sec.~\ref{s:analysis}). The XAI indicates the model's focal regions on the input image. Experts assess these areas for relevance and potential biases. By comparing these regions across images in the same category, common patterns are identified. These patterns are then cross-referenced with other categories to spot shared features. Additionally, we compare XAI results across various models to further address potential challenges.

\subsection{Solution Proposal}
The solution proposal phase is important for enhancing model performance. Once the issue is identified, experts review the dataset and model to identify potential causes like data distribution, labels, biases, or model design. Solutions may involve tweaking model parameters, refining training data, or enhancing the training procedure.

\subsection{Solution Assessment}
Rather than implementing all possible solutions, we shall assess the feasibility of proposed solutions on a small dataset initially. We evaluate the advantages and disadvantages of each solution, drawing from prior case studies to assess their relevance to the present problem. The infeasible solutions can be identified and eliminated, thereby allowing for the selection of the most suitable solution.

\subsection{Model Enhancement}
After implementing the effective solution identified earlier, we refine the model to address issues highlighted in Sec.~\ref{s:problem}. We then assess the model's enhancement by contrasting its performance pre and post-refinement, specifically comparing predictive metrics on initially selected images. Additionally, we might test using cases the original model struggled with to validate the model's enhanced capability in tackling the pinpointed issue.

\section{Experiment}
In our study, we detail each step as illustrated in Fig.~\ref{fig: xai process}. We experiment using the Bytetrack model pre-trained on datasets like MOT17 \cite{mot}, Cityperson \cite{cityperson}, ETHZ \cite{ethz}, and CrowdHuman \cite{crowdhuman}.

\subsection{Data Selection and Prediction Extraction}
Our training dataset amalgamates four public datasets~\cite{mot, cityperson, ethz, crowdhuman}.
We use CrowdHuman for our tests, divided into training (15000 images), validation (4370 images), and testing (5000 images) sets. These sets, with a combined 470K human instances, offer varied bounding box annotations. We choose a random 1000-image subset from CrowdHuman's training set for extracting model predictions, as outlined in Sec.~\ref{s:data_selection}.

\subsection{Analyzing Prediction Results}
Here, we match predicted boxes with the ground truth. ``Under-detection" is the predominant issue, constituting 85.5\%. While, ``Under-detection'' accounts for 17\%, ``Over-detection'' accounts for 10.8\%, and ``Mislocalization'' accounts for 20\%.

\begin{figure}[h]
    \centering
    \includegraphics[width=\linewidth]{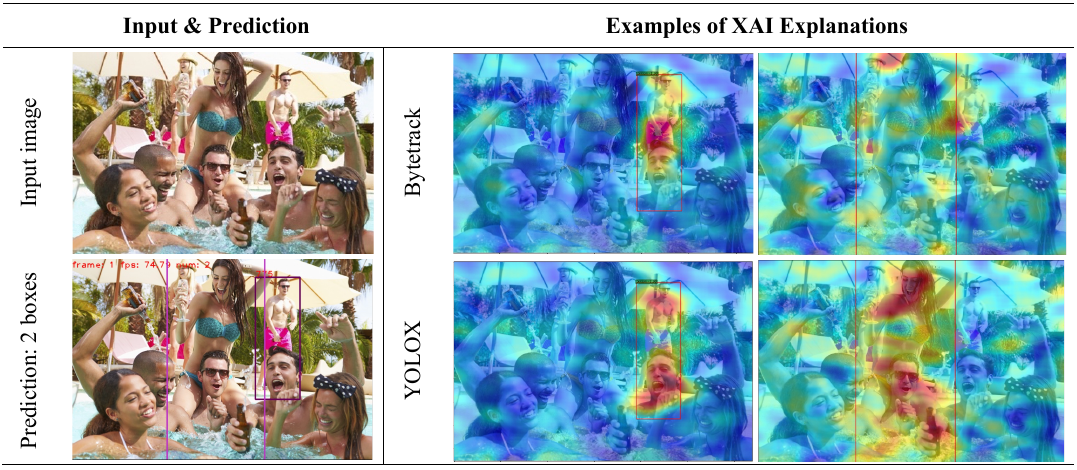}
    \caption{Examples of XAI Explanations with Bytetrack and YOLOX model. In which, each image in the second column is the XAI Explanations for a corresponding box.}\label{fig: xai bytetrack vs yolox}
    \vspace{-10pt}
\end{figure}

\subsection{Explanation Generation}
The Bytetrack model is a composite of YOLOX, responsible for detection, and the Byte phase that processes these detections. YOLOX is vital as the subsequent Byte step relies on its outputs. Byte's role is to maintain low-score predictions possibly hidden by other items~\cite{bytetrack}. We use D-RISE to interpret YOLOX, referencing the final box coordinates from Bytetrack~\cite{petsiuk2021black}. Additionally, comparing Bytetrack and YOLOX using D-RISE on YOLOX's weights aids in identifying differences, showcased in Fig.~\ref{fig: xai bytetrack vs yolox}~\cite{yolox}.

\subsection{Problem Identification}
The XAI explanations in Fig.~\ref{fig: xai bytetrack vs yolox} indicate Bytetrack's focus on entire human bodies, exposing its struggle to detect individuals showing only their heads. Experiments with images of people in wheelchairs, where bodies are partly concealed, amplify this limitation, with the model overlooking them as seen in Fig.~\ref{subfig:disabled}. Similar misses happen with people hidden behind objects, highlighted in Fig.~\ref{subfig:camera_cover_body}. Hence, Bytetrack's challenge in spotting partially visible humans emerges as a key concern needing attention and resolution.

\subsection{Solution Proposal}
\begin{figure}[h]
    \centering
    \includegraphics[width=\linewidth]{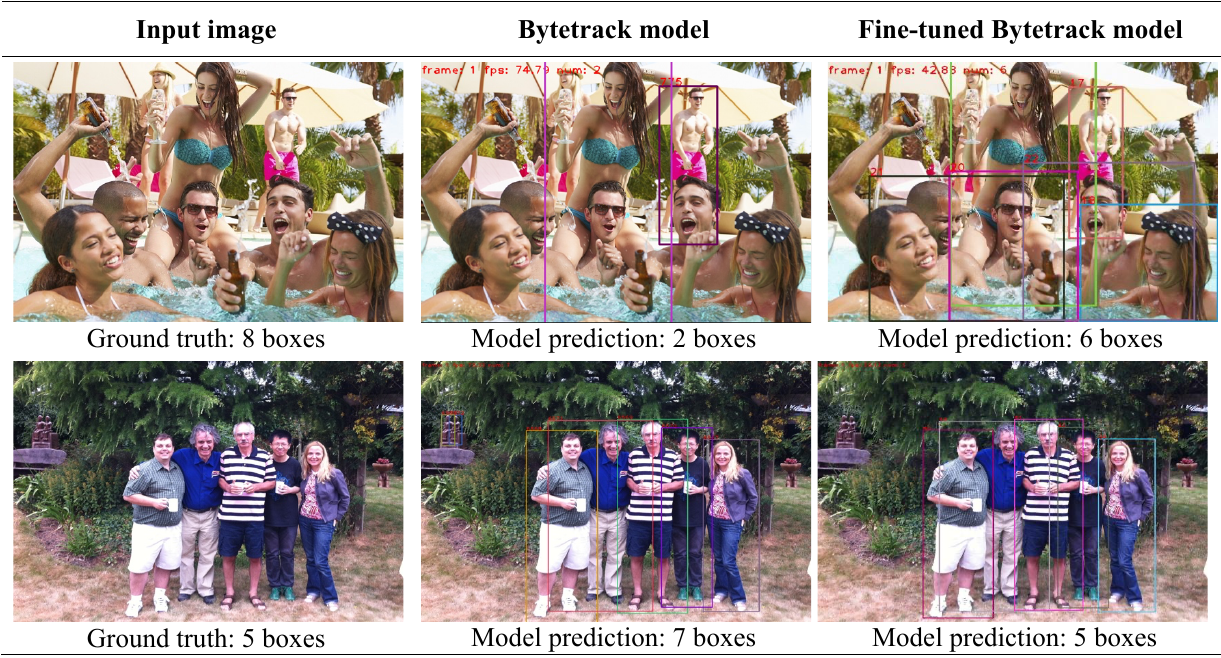}
    \caption{Predictions of the Bytetrack model before and after fine-tuning.}\label{tab: finetune result}
    \vspace{-10pt}
\end{figure}

We pinpointed specific issues and proposed assumptions accordingly:
\begin{itemize}
    \item Dataset: On average, images have 23 people, making heads smaller than bodies, potentially leading to a body bias. We also suspect label issues with ground truth box coordinates outside the image, shown in Fig.~\ref{fig: xai bytetrack vs yolox} and Table~\ref{tab: statistic result after fine-tune}.
    \item Model: Bytetrack tries to resolve occluded objects~\cite{bytetrack}. For head-only images, Bytetrack expects an associated body.
\end{itemize}

\begin{table}[h]
    \centering
    \caption{Ground truth boxes' coordinate of the input image in the first row of Fig.~\ref{fig: xai bytetrack vs yolox}, where 7/8 boxes are outside the image.}
    \resizebox{0.8\linewidth}{!}{%
    \begin{tabular}{ccccccccc}
    \toprule
    \textbf{Left} & -50 & -12 & 308 & 499 & 618 & 608 & 318 & 303 \\
    \textbf{Top} & 35 & 87 & 292 & 171 & 370 & 61 & -14 & -3 \\
    \textbf{Right} & 531 & 451 & 635 & 988 & 1034 & 758 & 673 & 444 \\
    \textbf{Bottom} & 131 & 1325 & 1228 & 1201 & 1243 & 444 & 745 & 437 \\
    \textbf{Outside image} & $\times$ & $\times$ & $\times$ & $\times$ & $\times$ & & $\times$ & $\times$ \\
    \bottomrule
    \end{tabular}}
    \vspace{-10pt}
\end{table}

Proposed solutions include:
\begin{itemize}
    \item Data enrichment: Add images with mostly obscured body sections.
    \item Data blurring: Based on XAI findings, blur bodies to make the model focus on heads.
    \item Padding: Ensure bounding boxes are fully within images.
    \item Relabeling: Adjust bounding boxes to remain inside the image.
\end{itemize}

\subsection{Solution Assessment}
\begin{figure}[h]
    \centering
    \includegraphics[width=\linewidth]{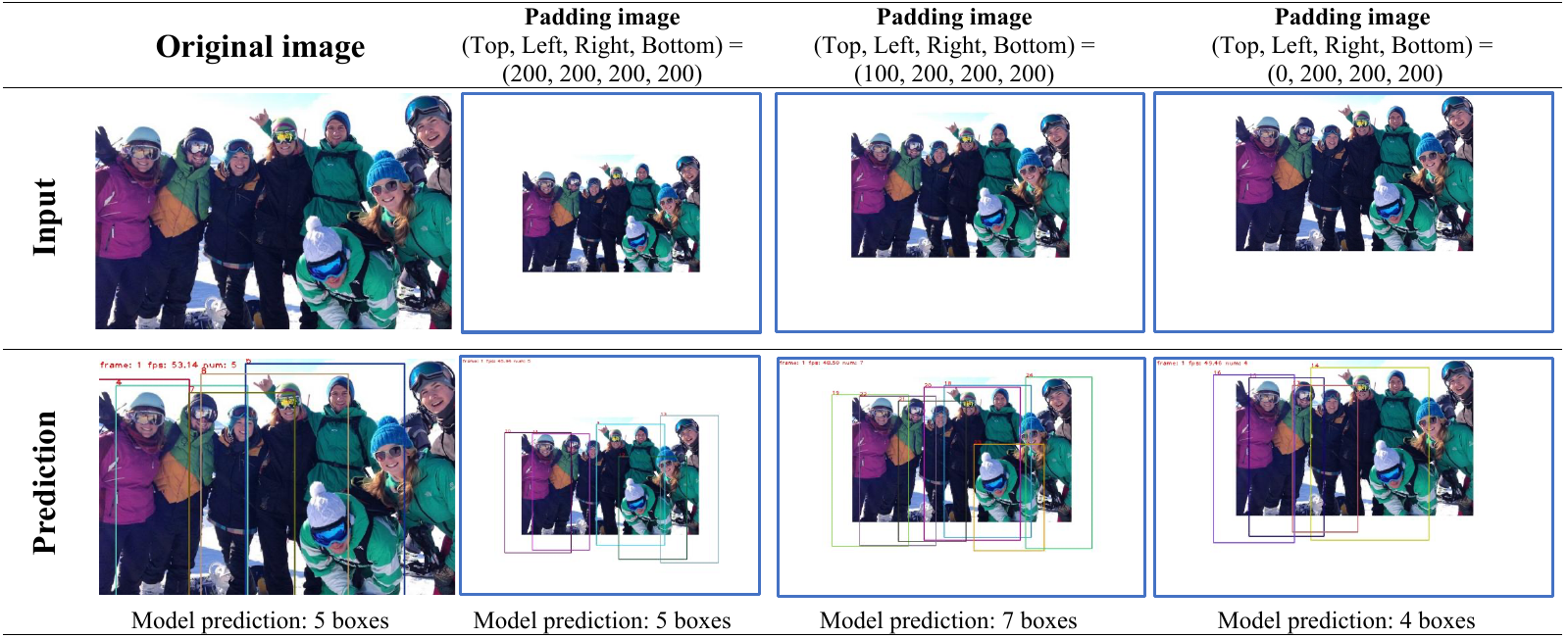}
    \caption{Example of padding result. (Top, Left, Right, Bottom) = (100, 200, 200, 200) signifies padding of 100, 200, 200, and 200 pixels respectively on the top, left, right, and bottom.}\label{fig: padding example}
    \vspace{-10pt}
\end{figure}

We conduct a comprehensive analysis to identify and implement the most suitable solution to the problem. Each solution is evaluated as follows:
\begin{itemize}
\item Data enrichment: The current dataset already has partly hidden figures, so more data might not help much.
\item Data blurring: Effective for image classification, but might not suit human detection where only humans are predicted.
\item Padding: While sometimes effective, as in Fig.~\ref{fig: padding example}, it often fails, especially when objects obstruct people.
\item Relabeling: Given dataset inconsistencies and variant model features, relabeling seems promising.
\end{itemize}
Following this analysis, relabeling emerges as the most impactful solution.

\subsection{Model and Dataset Enhancement}    

\begin{figure}
    \centering
    \resizebox{0.8\linewidth}{!}{%
    \begin{tabular}{ccccc}
       \toprule
        Bytetrack model & Fine-tuned Bytetrack model \\
        \midrule
        \includegraphics[width=3cm]{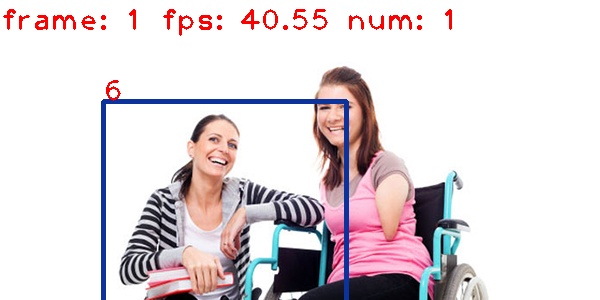} & \includegraphics[width=3cm]{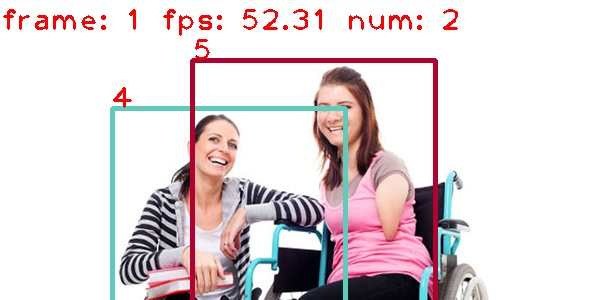} \\
        \includegraphics[width=2.5cm]{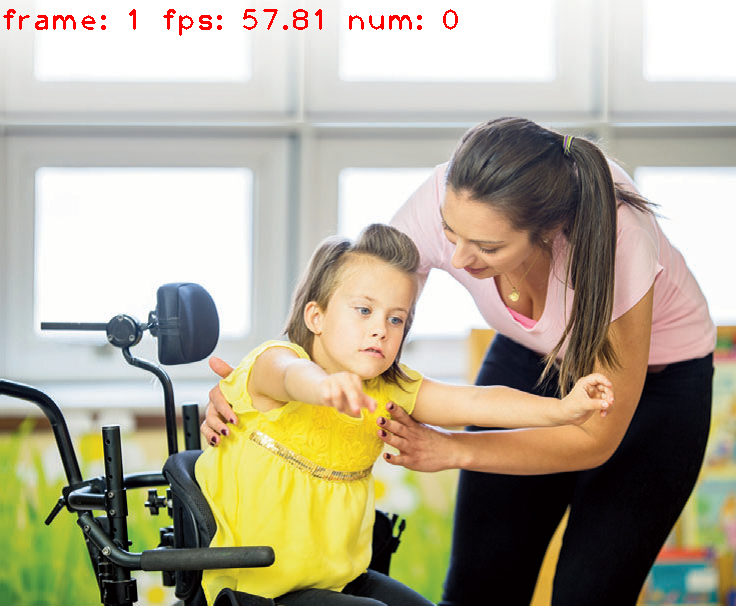} & \includegraphics[width=2.5cm]{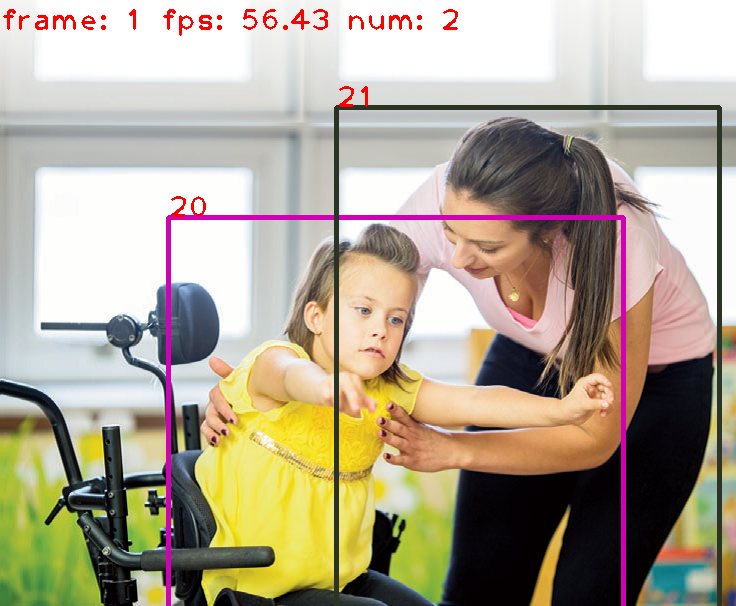} \\
        \bottomrule
    \end{tabular}}
    \caption{Model's prediction on physically disabled person images. After fine-tuning, the model performs better than the original pre-trained model.}   \label{table:dissability_after_finetune}
    \vspace{-10pt}
\end{figure}

\begin{figure}[h]
    \centering
    \includegraphics[width=.9\linewidth]{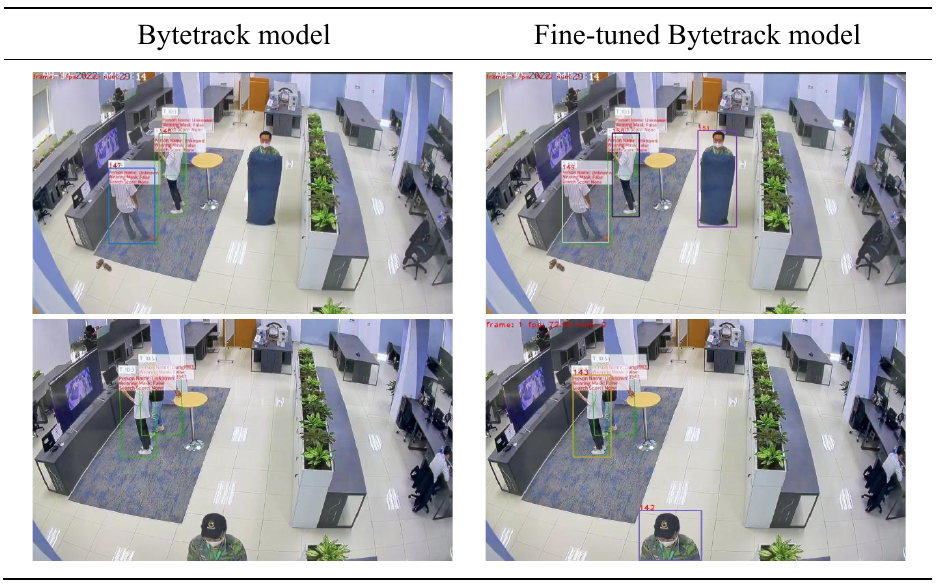}
    \caption{Model's prediction on a security camera. The fine-tuned model performs better than the original pre-trained model detecting covered people.}\label{fig: camera test model after fine-tune}
    \vspace{-10pt}
\end{figure}
        
The CrowdHuman dataset is reannotated by constraining bounding box coordinates within the image dimensions, as delineated by $x^{'}_{\text{top, left}} = \texttt{max}(0, x_{\text{top, left}})$, $y^{'}_{\text{top, left}} = \texttt{max}(0, y_{\text{top, left}})$, $x^{'}_{\text{bottom, right}} = \texttt{min}(w, x_{\text{bottom, right}})$, $y^{'}_{\text{bottom, right}} = \texttt{min}(h, y_{\text{bottom, right}})$.
Here, $w, h$ represents the image's width and height, respectively. The coordinates $(x^{'}_{\text{top, left}}, y^{'}_{\text{top, left}})$ and $(x^{'}_{\text{bottom, right}}, y^{'}_{\text{bottom, right}})$ denote the adjusted top-left and bottom-right points, respectively. Subsequent model refinement occurs over 10 epochs, with performance enhancement evaluated in three scenarios:

\begin{itemize}
    \item Training Dataset Testing: We test a 1000-image subset after refining the model. Both quantitative and qualitative evaluations are made against the original model, as seen in Table \ref{tab: statistic result after fine-tune} and Fig.~\ref{tab: finetune result}. The updated model better localizes in 855 ``Under-detection" images, improving by 21 cases.

    \item Images of Disabled Individuals: The adjusted model shows better detection in images featuring physically disabled people, highlighted in Fig.~\ref{table:dissability_after_finetune}.
    
    \item Detection in Surveillance Footage: We assess the model in real-life contexts, like office security footage where people might be partly hidden. Post-refinement performance, showcasing improvements, is depicted in Fig.~\ref{fig: camera test model after fine-tune}.
\end{itemize}

 \begin{table}[h]
        \centering
        \caption{Statistical result pre-trained model versus fine-tuned model. The arrow $\uparrow$/$\downarrow$ indicates the higher/lower value, the better. The bold indicates the better result.}
        \resizebox{.8\linewidth}{!}{%
            \begin{tabular}{ccc} 
            \toprule
            Case & Pre-trained model & Fine-tuned model \\ 
            \midrule
            Under-detection ($\downarrow$)        & 855  &      \textbf{834}     \\
            
            Over-detection ($\downarrow$)      & 17   &      \textbf{13}       \\ 
            
            Correct Localization ($\uparrow$)         & 108   &       \textbf{133}      \\ 
            
            Mislocalization ($\downarrow$)       & 20     &      20     \\ 
            \bottomrule
            \end{tabular}
            }
        \label{tab: statistic result after fine-tune}
        \vspace{-12pt}
    \end{table}

\section{Conclusion and Future work}
This study introduces a human detection debugging framework using XAI aided by experts. Our approach pinpoints data labeling as a significant issue in Bytetrack's biases and can adapt to other detection problems, especially those focusing on specific classes.

\bibliographystyle{IEEEtran}
\bibliography{IEEEabrv,ref}

\end{document}